# Deep Learning-based Fall Detection Algorithm Using Ensemble Model of Coarse-fine CNN and GRU Networks

Chien-Pin Liu
*Department of Biomedical Engineering*
*National Yang Ming Chiao Tung University*
Taipei City 112, Taiwan
henry062439.be09@nycu.edu.tw

Ju-Hsuan Li
*Department of Biomedical Engineering*
*National Yang Ming Chiao Tung University*
Taipei City 112, Taiwan
yeach52639.y@nycu.edu.tw

En-Ping Chu
*Department of Biomedical Engineering*
*National Yang Ming Chiao Tung University*
Taipei City 112, Taiwan
robinchu.be07@nycu.edu.tw

Chia-Yeh Hsieh
*Bachelor's Program in Medical Informatics and Innovative Applications*
*Fu Jen Catholic University*
New Taipei City 242062, Taiwan
152369@mail.fju.edu.tw

Kai-Chun Liu
*Research Center for Information Technology Innovation*
*Academia Sinica*
Taipei City 115, Taiwan
t22302856@citi.sinica.edu.tw

Chia-Tai Chan
*Department of Biomedical Engineering*
*National Yang Ming Chiao Tung University*
Taipei City 112, Taiwan
ctchan@nycu.edu.tw

Yu Tsao
*Research Center for Information Technology Innovation*
*Academia Sinica*
Taipei City 115, Taiwan
yu.tsao@citi.sinica.edu.tw

*Abstract*—Falls are the public health issue for the elderly all over the world since the fall-induced injuries are associated with a large amount of healthcare cost. Falls can cause serious injuries, even leading to death if the elderly suffers a "long-lie." Hence, a reliable fall detection (FD) system is required to provide an emergency alarm for first aid. Due to the advances in wearable device technology and artificial intelligence, some fall detection systems have been developed using machine learning and deep learning methods to analyze the signal collected from accelerometer and gyroscopes. In order to achieve better fall detection performance, an ensemble model that combines a coarse-fine convolutional neural network and gated recurrent unit is proposed in this study. The parallel structure design used in this model restores the different grains of spatial characteristics and capture temporal dependencies for feature representation. This study applies the FallAllD public dataset to validate the reliability of the proposed model, which achieves a recall, precision, and F-score of 92.54%, 96.13%, and 94.26%, respectively. The results demonstrate the reliability of the proposed ensemble model in discriminating falls from daily living activities and its superior performance compared to the state-of-the-art convolutional neural network long short-term memory (CNN-LSTM) for FD.

## I. Introduction

Falls are a common health issue worldwide, particularly for the elderly. Globally, one-third of people aged over 65 fall at least once per year [1]. Poor physical ability and fitness increase the chance of injury and death when a fall occurs [2, 3]. In addition, the association between the mortality rate and waiting time for rescue shows a positive correlation [4]. When the elderly experience a fall and lie on the ground for more than 2.5 h (a "long lie"), the mortality rate increases to 50% [5]. Therefore, developing a reliable fall detection (FD) system for first aid purposes is essential.

With the progress of micro-electro-mechanical systems (MEMS), several types of wearable sensors have been used to collect long-term sequential data for FD systems, including wearable accelerometers [6-10], barometers [9], and cameras [11]. Among these, wearable accelerometers are the most popular for collecting data in FD system development because of their low cost, compactness, miniaturization, and noninvasive nature. Previous studies [6-10] have demonstrated the feasibility of accelerometer-based FD systems.

Several studies have developed FD systems using typical machine learning (ML) methods, such as support vector machine (SVM) [4, 6, 7], Naïve Bayes (NB) [6], k-nearest neighbor (kNN) [6], and decision tree (DT) [6]. These methods require several handcrafted features, relying on the domain knowledge of researchers and requiring extensive research. A previous study demonstrated the limited detection performance of typical ML methods in more complicated experimental setups [9]. Compared with typical ML, deep learning (DL) methods, including convolutional neural network (CNN) and recurrent neural network (RNN), can extract features automatically and directly from raw data. Several studies have shown that DL-based FD systems have better detection abilities than typical ML-based approaches [9, 12, 13].

CNNs are generally applied in FD systems [9, 13], which can extract the spatial features of raw data using convolutional filters. Santos et al. [10] proposed a CNN-based FD system with data augmentation preprocessing that achieved 99.86%,

100%, and 99.72% in accuracy, precision, and sensitivity, respectively. RNNs, which are another common DL model that use cells to store time-series information, are ideal for analyzing temporal data. As a fall can be represented as a temporal sequence, RNNs are used in FD systems to address the temporal characteristics. Two common variants, the long short-term memory (LSTM) and gated recurrent unit (GRU), were developed to avoid the vanishing gradient problem in the traditional RNN model. Francisco et al. [7] validated the effectiveness of RNN in wearable-based FD systems. Each temporary segment was classified as a fall event, fall hazard, or activity of daily living (ADL). Their method achieved 96.7%, 69.5%, 90.2%, 73.0% in accuracy, precision, sensitivity, and F-score, respectively. They also demonstrated that a GRU-based FD system has a detection performance similar to that of an LSTM-based FD system but requires a slightly lower computing complexity owing to the smaller number of parameters.

Some research has focused on combining CNN and RNN in different structures to further improve FD performance. Xu et al. [14] developed a state-of-the-art model for FD called CNN-LSTM, which applies a sequential structure that passes the features extracted by convolutional filters to the RNN layers in series. Their results demonstrated that FD using CNN-LSTM has a better detection ability than conventional RNN. However, combining a CNN and RNN in a sequential structure may dilute the contribution of the CNN features to the FD ability, whereas the RNN post-processes features.

To address this issue, a parallel structure may serve as a viable approach to reserve critical CNN features for FD. For example, Avilés-Cruz et al. [15] combined CNN networks with a parallel structure named coarse-fine CNN for ADLs, involving several branches with an inconsistent number of convolution and max-pooling layers to reserve different grains of features and provide more information for training a better classifier. The coarse-fine CNN method used in [15] showed superior performance compared to typical CNN methods. However, such a parallel structure combining the coarse-fine CNN model has not yet been applied in FD systems.

Motivated by previous approaches, we propose an ensemble neural network model of a parallel structure combining a coarse-fine CNN and GRU for an FD system. The proposed method applies a coarse-fine CNN and GRU to gather the different grains of spatial and temporal features, respectively. A parallel structure can restore the features extracted by different neural networks and avoid possible dilution in a sequential structure. In addition, the proposed model employs a simple concatenation process to merge the extracted spatial and temporal features. We conducted experiments on the public FallAllD [9] dataset to validate the reliability of the proposed framework. The results show that the proposed network successfully incorporates temporal and coarse-fine spatial features and achieves superior performance compared to other DL models for FD.

## II. II. MATERIAL AND METHODS

### A. Public FallAllD Dataset

FallAllD [9] is a public dataset containing several types of falls and ADLs recorded by inertial measurement units (IMUs) worn on the neck, waist, and wrist of subjects. A total of 6605 instances, including 1722 fall instances and 4883 ADL instances, were performed by 15 subjects (eight males and seven females). Each instance contained streaming data for 20 s. The transition phase of the ADL instances and the impact point of the fall instances were centered at the 10th second of each instance. Each IMU consisted of a tri-axial accelerometer (measuring range: ±8 g; sampling rate: 238 Hz), a tri-axial gyroscope (measuring range: ±2000 degree per second; sampling rate: 238 Hz), a tri-axial magnetometer (measuring range: ±4 Gauss; sampling rate: 80 Hz), and a barometer (measuring range: 10-1200 mbar; sampling rate: 10 Hz).

Only tri-axial acceleration data recorded from the IMU on the waist was applied in this study to validate the proposed ensemble FD network, which involved 466 fall instances and 1332 ADL instances. The preprocessing process adopts downsampling and sliding-window techniques to segment raw data as input for the proposed model. Each instance was downsampled from 238 Hz to 20 Hz. The sliding window technique, with a window size of 7 s and 50% overlapping, was applied to divide each instance. For each ADL instance, all segments were adopted as the data, but only segments containing the fall impact signal were adopted as fall instances. A total of 6260 segments were used as input data through preprocessing, including 5328 ADL segments and 932 fall segments. Each segment is a two-dimensional data point with a size of 3 × 140.

### B. CNN

A typical CNN comprises convolution layers, a nonlinear activation function, pooling layers, flattening, and fully connected layers. The convolution layers extract spatial features from the raw data with multiple filters sliding over the input data. The summation of the element-wise multiplication of a filter and receptive field of the input data is calculated as the layer output, as shown in Fig. 1. Each filter is trained using backpropagation and can represent different aspects of the input data. Nonlinear activation functions, such as the rectified linear unit (ReLU), sigmoid (S), and hyperbolic tangent (tanh) functions, are applied to the outputs of the convolution and fully connected layers to improve the accuracy of the classifier. The ReLU function, expressed by Equation (1), is commonly used in CNN

$$ReLU(x) = \begin{cases} 0, & x < 0 \\ x, & x \geq 0 \end{cases} \quad (1)$$

Pooling layers reduce the dimensions of the input data by applying certain rules inside the filter. Max pooling, the most common method, retrieves the maximum value inside the filter as output. After the convolution and pooling layers, the extracted two-dimensional features are flattened into one-dimensional representations to be input into the fully connected layers. A fully connected layer computes the output vector using Equation (2), as follows:

$$y = \sigma(wx + b) \quad (2)$$

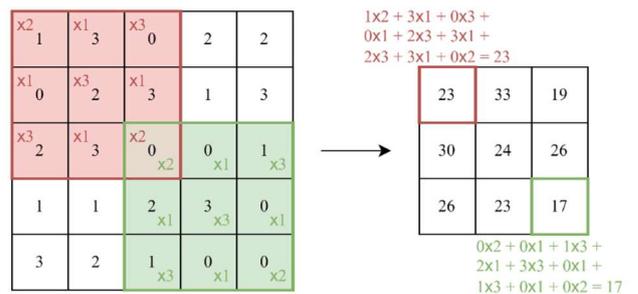

Fig. 1 Illustration of convolutional computation

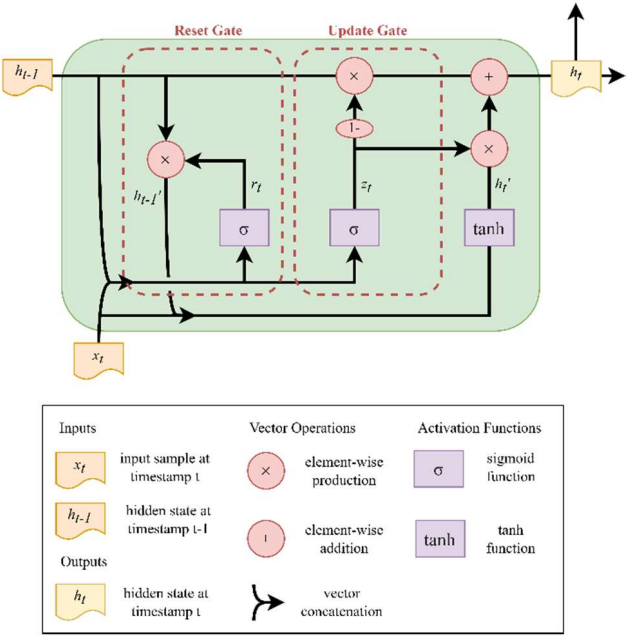

Fig. 2 Structure of a GRU component

where $y$ is the output vector, $x$ the input vector, $w$ the weight matrix, $b$ the bias vector, and $\sigma$ is the layer activation function. We used the softmax function for the last fully connected layer and the ReLU function for the other fully connected layers. The softmax function was implemented to produce the normalized probabilities for each class. The coarse-fine CNN has the same structure as a typical CNN except for two branches with an inconsistent number of convolutional and max-pooling layers.

*C. GRU*

A GRU network is a variant of an RNN that significantly reduces the influence of the vanishing gradient problem, with similar performance but lower computing complexity than an LSTM network. A GRU network typically comprises GRU layers for extracting temporal features and fully connected layers to use the features for classification. Fig. 2 shows the structure of a GRU component. The $t^{th}$ GRU component outputs hidden state $h_t$, which is a vector that stores the information of past and present inputs, with input value $x_t$ and hidden state $h_{t-1}$ produced by the last GRU component. The variables in Fig.2 were calculated using Equations (3)-(7). In these equations, parameters $W_{rt}, W_{zt}, W_t, b_{rt}, b_{zt}$, and $b_t$ were trained by backpropagation, where $W$ is the weight matrix, and $b$ the bias vector. $r_t$ is the parameter of the reset gate, and $z_t$ is the parameter of the update gate, whose range is (0,1). The greater the $z_t$, the more information $h_t$ stores about $x_t$. Operator × and + represent element-wise production and addition, respectively. Two activation functions, S and tanh, were applied; these are presented as Equations (8)–(9).

$$r_t = S(W_{rt}[h_{t-1}, x_t] + b_{rt}) \quad (3)$$

$$z_t = S(W_{zt}[h_{t-1}, x_t] + b_{zt}) \quad (4)$$

$$\tilde{h}_{t-1} = r_t \times h_{t-1} \quad (5)$$

$$\tilde{h}_t = \tanh\left(W_t\left[[\tilde{h}_{t-1}, x_t] + b_t\right]\right) \quad (6)$$

$$h_t = h_{t-1} \times (1 - z_t) + \tilde{h}_t \times z_t \quad (7)$$

$$S(x) = \frac{1}{1 + e^{-x}} \quad (8)$$

$$\tanh(x) = \frac{e^x - e^{-x}}{e^x + e^{-x}} \quad (9)$$

*D. Proposed Model*

The proposed ensemble model comprises three branches for extracting different motion characteristics from the raw data: the coarse, fine, and temporal branches. Each branch contains a flattening procedure at the end to reshape the extracted features from two dimensions to one. Figure 3 illustrates the structure of the proposed ensemble model. The three branches were designed to extract spatial and temporal features from the input sequence.

- Coarse branch: The first branch extracts coarse-grained spatial features from the raw signal. It comprises one convolutional layer with 32 3 × 3 filters and one max pooling layer with a 1 × 2 filter.

- Fine branch: The second branch extracts fine-grained spatial features from the accelerometer signals. It involves two groups of a convolution layer with 32 filters and a max-pooling layer with a 1 × 2 filter. The convolution layer filter size was 3 × 3 in the first group and 1 × 3 in the second.

- Temporal branch. This branch captures temporal and contextual information from the movement signals performed. This model has two GRU layers, and each layer uses a 64-element array to represent the hidden state at each timestamp.

The outputs of each branch are concatenated into a one-dimensional vector as input for the fully connected layer. The first and second fully connected layers had 64 and two neurons, respectively. Softmax was the activation function of all convolutional layers and the first fully connected layer.

The total loss of the system is the summation of the loss functions of the three branches, as expressed in Equation (10).

$$loss_{total} = loss_{coarse} + loss_{fine} + loss_{temporal} \quad (10)$$

All loss functions are binary cross-entropies. The optimizer was Adam, with an initial learning rate of 0.01. Forty epochs were applied for model training, and the batch size for each epoch was 32. The proposed ensemble model was processed and analyzed on Python 3.8 in a Windows 10 environment. For training, we used an NVIDIA GTX 1060 graphic processor unit, and the CuDNN versions of the RNN layer provided were implemented in the Keras framework.

*E. Performance Evaluation*

This study applies the leave-one-subject-out (LOSO) cross-validation method to evaluate the performance of the proposed model. This method uses the data of one subject as the testing set and the remaining data as the training set. The process was repeated until all subject data were tested once.

Three common evaluation metrics were considered to assess the model: accuracy, recall, precision, and F-score. The metrics were computed using Equations (11)–(14):

$$accuracy = \frac{TP + TN}{TP + FP + FN + TN} \quad (11)$$

Table 1 Comparison table of performances of different models

| Method | Evaluation Metrics | | | |
|---|---|---|---|---|
| | Accuracy (%) | Recall (%) | Precision (%) | F-score (%) |
| Simple CNN | 97.82 | 90.88 | 95.77 | 93.52 |
| Simple GRU | 97.49 | 91.60 | 94.01 | 92.71 |
| Coarse-fine CNN | 97.78 | 91.82 | 95.43 | 93.52 |
| CNN-LSTM | 97.90 | 90.78 | 95.49 | 93.48 |
| CNN-GRU | 97.69 | 89.56 | 95.42 | 92.24 |
| The proposed model | **97.95** | **92.54** | **96.13** | **94.26** |

$$recall = \frac{TP}{TP + FN} \quad (12)$$

$$precision = \frac{TP}{TP + FP} \quad (13)$$

$$F - score = 2 \times \frac{recall \times precision}{recall + precision} \quad (14)$$

where TP, FN, and FP are the true positive, false negative, and false positive, respectively. As the primary purpose of the proposed model is to detect fall events, falls and ADLs were defined as positive and negative, respectively.

### III. RESULT AND DISCUSSION

Several typical models were implemented in FD systems using the FallAllD dataset to validate the superiority of the proposed ensemble model. For fairness, all models followed the same experimental setup as the proposed method, with their parameters optimized. Each structure was set up as follows:

Simple CNN: Adopting a CNN architecture comprising two convolutional layers and two pooling layers.

Simple GRU: Utilizing two GRU layers to process raw sensor data.

Coarse-fine CNN: The coarse branch of Coarse-fine CNN uses a one-layer CNN structure; the fine branch shares the same architectural design as Simple CNN.

CNN-LSTM: The state-of-the-art model for FD shares the same architectural design and hyperparameters as in a previous study [14]. It consists of two convolution layers, followed by two LSTM layers.

CNN-GRU: The LSTM layers in the CNN-LSTM were replaced with GRU layers because these were used in the proposed model to extract temporal features. The remaining structure and hyperparameters were the same as that with CNN-LSTM.

As shown in Table 1, the proposed model achieved recall, precision, and F-score values of 92.54%, 96.13%, and 94.26%, respectively. The CNN and Coarse-Fine CNN models have the best F-score of 93.52 % for FD among the other DL methods, revealing that FD using the proposed ensemble model outperforms those using typical DL approaches. In addition, the FD model using two-layer GRU achieves an F-score of 92.71%, slightly lower than that of the CNN-based methods (93.52%). The performance of FD using these typical DL models are similar to that in a previous study [9].

The state-of-the-art FD model developed by Xu et al. [14] has been widely used to compare FD tasks [14, 16, 17]. To validate the superiority of the proposed ensemble approach, the CNN-LSTM-based FD is tested on the FallAllD dataset for comparison purposes. The results demonstrate that FD system using CNN-LSTM achieve F-score, recall, and precision values of 93.48%, 90.78%, and 95.49%, respectively; the proposed ensemble approach improves the F-score, recall, and precision of FD by 1.76%, 0.64%, and 0.74%, respectively.

To the best of our knowledge, the proposed approach is the first to apply a parallel structure design to increase FD performance. We validate the design using a parallel structure that could provide FD system with better detection abilities than using a sequential structure combining CNN and LSTM. An ensemble model involving coarse-fine CNN and GRU branches was proposed to further enhance the FD ability. The results reveal that a parallel structure design that restores CNN and RNN features benefits FD, however, passing the output of convolutional layers to RNN can lose critical spatial characteristics in typical sequential structures (e.g., CNN-LSTM and CNN-GRU). The feasibility of the parallel structure has also been validated in other classification studies [18]. In addition, a previous study [7] demonstrated that an FD system using GRUs can reduce computational complexity and achieve a detection performance similar to that using LSTM layers. Therefore, GRU layers, instead of LSTM layers, were employed in the proposed ensemble FD model. This setup allows the proposed FD approach to support the creation of a wearable-based FD system for long-term healthcare services.

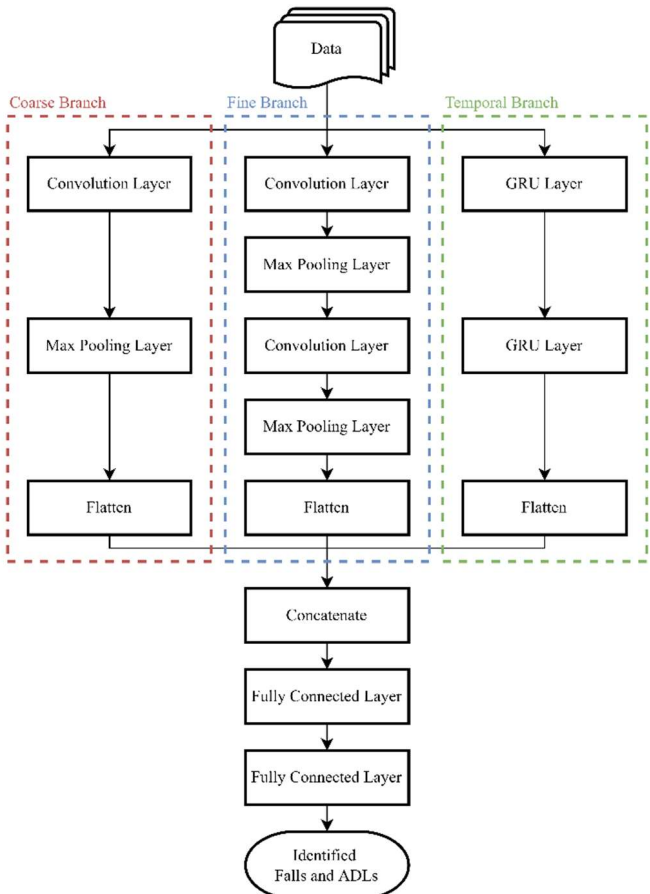

Fig. 3 Framework of the proposed ensemble model

## IV. Conclusion

Falls are a concerning health issue among the elderly worldwide. To address this issue, a reliable and automatic FD system is required to detect falls and alert emergency services. This study proposes an ensemble model that combines a coarse-fine CNN and GRU to extract spatial and temporal features for FD. We show that FD using the proposed model achieves 92.54% in recall, 96.13% in precision, and 94.26% in the F-score and achieves superior performance to using typical DL models and the state-of-the-art CNN-LSTM. The results show that the proposed method can provide an FD system with a better ability to discriminate falls from ADLs.

Future work will involve additional datasets to validate the effectiveness of the proposed approach. Several advanced models and mechanisms, such as residual learning [19] and self-attention [20] mechanisms, will be incorporated to enhance detection performance.

Acknowledgment

This work was supported in part by grants from the Ministry of Science and Technology.